\documentclass[sigconf]{acmart}
\AtBeginDocument{%
  }

\setcopyright{acmlicensed}
\copyrightyear{2026}
\acmYear{2026}
\acmDOI{XXXXXXX.XXXXXXX}
\acmConference[HRI '26]{International Conference on Human-Robot Interaction}{Mar 16--19, 2026}{Edinburgh, UK}
\acmISBN{978-1-4503-XXXX-X/2018/06}

\usepackage{tabularx,booktabs}
\usepackage{enumitem}
\usepackage[per-mode=symbol]{siunitx}
\usepackage{color}

\usepackage{lscape}
\usepackage{pifont}
\usepackage{xcolor}
\usepackage{makecell}
\usepackage{soul}

\usepackage{multirow}
\usepackage{graphicx} 

\usepackage{booktabs}  
\usepackage{xcolor} 

 \usepackage{pifont}

\hypersetup{
    colorlinks=true,
    linkcolor=blue,
    filecolor=magenta,      
    urlcolor=magenta,
    pdftitle={Overleaf Example},
    pdfpagemode=FullScreen,
    }

\usepackage{tabularx,booktabs}
\newcolumntype{Y}{>{\raggedright\arraybackslash}X}

\begin{document}


\title{\textsc{GuideNav}: User-Informed Development of a Vision-Only \\Robotic Navigation Assistant For Blind Travelers}

\author{Hochul Hwang}
\email{hochulhwang@umass.edu}
\orcid{0000-0002-3199-7208} 
\affiliation{%
  \institution{University of Massachusetts Amherst}
  \city{Amherst}
  \state{Massachusetts}
  \country{USA}
}

\author{Soowan Yang}
\email{tntn6771@dgist.ac.kr}
\orcid{0009-0003-9903-2781}
\affiliation{%
  \institution{DGIST}
  \city{Daegu}
  \country{Republic of Korea}
}

\author{Jahir Sadik Monon}
\email{jmonon@umass.edu}
\orcid{0009-0006-9384-596X} 
\affiliation{%
  \institution{University of Massachusetts Amherst}
  \city{Amherst}
  \state{Massachusetts}
  \country{USA}
}

\author{Nicholas A Giudice}
\email{nicholas.giudice@maine.edu}
\orcid{0000-0002-7640-0428} 
\affiliation{%
  \institution{University of Maine}
  \city{Orono}
  \state{Maine}
  \country{USA}
}

\author{Sunghoon Ivan Lee}
\email{silee@cs.umass.edu}
\orcid{0000-0001-5935-125X} \affiliation{%
  \institution{University of Massachusetts Amherst}
  \city{Amherst}
  \state{Massachusetts}
  \country{USA}
}

\author{Joydeep Biswas}
\email{joydeepb@cs.utexas.edu}
\orcid{0000-0002-1211-1731} 
\affiliation{%
  \institution{University of Texas at Austin}
  \city{Austin}
  \state{Texas}
  \country{USA}
}

\author{Donghyun Kim}
\email{donghyunkim@umass.edu}
\orcid{0000-0001-9534-5383} 
\affiliation{%
  \institution{University of Massachusetts Amherst}
  \city{Amherst}
  \state{Massachusetts}
  \country{USA}
}

\renewcommand{\shortauthors}{Hwang et al.}






\begin{abstract}
While commendable progress has been made in user-centric research on mobile assistive systems for blind and low-vision (BLV) individuals, references that directly inform robot navigation design remain rare. To bridge this gap, we conducted a comprehensive human study involving interviews with 26 guide dog handlers, four white cane users, nine guide dog trainers, and one O\&M trainer, along with 15+ hours of observing guide dog–assisted walking. After de-identification, we open-sourced the dataset to promote human-centered development and informed decision-making for assistive systems for BLV people. Building on insights from this formative study, we developed \textsc{GuideNav}, a vision-only, teach-and-repeat navigation system. Inspired by how guide dogs are trained and assist their handlers, \textsc{GuideNav} autonomously repeats a path demonstrated by a sighted person using a robot. Specifically, the system constructs a topological representation of the taught route, integrates visual place recognition with temporal filtering, and employs a relative pose estimator to compute navigation actions---all without relying on costly, heavy, power-hungry sensors such as LiDAR. In field tests, \textsc{GuideNav} consistently achieved kilometer-scale route following across five outdoor environments, maintaining reliability despite noticeable scene variations between teach and repeat runs. A user study with 3 guide dog handlers and 1 guide dog trainer further confirmed the system’s feasibility, marking (to our knowledge) the first demonstration of a quadruped mobile system retrieving a path in a manner comparable to guide dogs.
\end{abstract}


\begin{teaserfigure}
  \includegraphics[width=\textwidth]{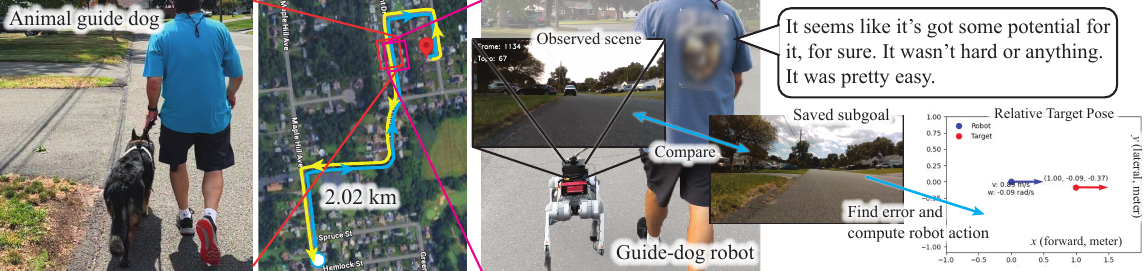}
  \caption{We developed a vision-only route following system considering technical requirements for practical guide robots. Our guide dog robot achieved kilometer-scale guiding with blind and low-vision (BLV) users. Robust path following performance and positive feedback from BLV users mark an important milestone toward the real-world deployment of guide dog robots.}
  \label{fig:teaser}
  \Description{Four-panel figure showing guide dog robot system. Panel 1: Person with guide dog walking on a sidewalk. Panel 2: Aerial view of route with yellow and blue path overlay showing 2.02 km distance and colorful bounding boxes tracking the person. Panel 3: Split view showing robot-perspective observed street scene at top and subgoal at bottom, with arrows indicating comparison between saved subgoal image and current view, plus a plot showing relative target pose with forward distance in meters. Panel 4: Quote bubble stating ``It seems like it's got some potential for it, for sure. It wasn't hard or anything. It was pretty easy.'' with an arrow pointing to the process of finding the error and computing the robot action.}
\end{teaserfigure}




\maketitle






\section{Introduction}
What should be the first step in developing a mobile navigation assistant system for blind and low-vision (BLV) individuals? The answer would be to identify the most meaningful and important needs of BLV travelers. While commendable strides have been made in BLV user-centered development of navigation systems \cite{guerreiro2019cabot, kuribayashi2023pathfinder, kamikubo2025beyond, ranganeni2023exploring, takagi2025field, kuribayashi2025wanderguide, slade2021multimodal, hwang2024towards,cai2024navigating}, defining a navigation system that is truly practical for BLV individuals remains challenging. Building a navigation assistant system for BLV travelers is not simply a matter of replicating autonomous driving systems---like a delivery robot using dense map-based navigation moving from any point A to B on a prebuilt map~\cite{starship}. Prior studies on BLV navigation assistance showed that the level of navigation assistance can vary widely, from mobility-only support (i.e., obstacle avoidance) to full autonomy (i.e., governing navigation direction)~\cite{ranganeni2023exploring}. 
Other works~\cite{takagi2025field, kuribayashi2025wanderguide} investigated exploratory navigation setups that allow BLV travelers to visit new places, such as exhibitions, with an additional scene description assistance. Studies on interactive navigation of guide dogs and handlers have also highlighted the importance of trust and a sense of agency as prerequisites for introducing autonomous navigation to BLV people~\cite{hwang2024towards}.

Such prior research provides valuable insights into BLV navigation relating to robotic guides; however, many detailed technical questions remain unanswered: What is the proper sensor suite---can LiDAR be an appropriate option? What conditions (e.g., terrain types, weather, lighting) must the robot handle? How much distance should the system support for daily navigation? And is it valid to assume that users have prior knowledge of the places they navigate? Answering such questions requires multi-dimensional considerations. For example, evaluating the suitability of LiDAR depends on power consumption, onboard computing capacity, operating conditions, and cost. These factors, in turn, are shaped by BLV users’ daily navigation patterns: how far they typically walk, whether they commute in rain, fog, or snow, and the acceptable robot weight (which constrains battery and computer size).

These questions may never be fully resolved until a system is deployed to end users. Despite extensive research on spatial cognition and navigation in psychology, cognitive science~\cite{Epstein2017}, and HRI, resources capturing the lived practices of BLV travelers remain scarce. This constrains the translation of theoretical insights into practical, human-centered navigation systems. By examining BLV users’ daily activities, commuting habits, common destinations, and the shortcomings of current mobility aids, development errors can be significantly reduced. Unfortunately, open-source data capturing this information is minimal, creating a major obstacle for human-centered research on mobile navigation systems. 

To bridge this gap, we conducted a comprehensive formative study involving interviews with 26 guide dog handlers, four cane users, nine guide dog trainers, and one O\&M trainer, along with 15+ hours of observing how guide dog handlers navigate with their dogs. We specifically examined how BLV individuals interact with guide dogs, which informed the design of our pragmatic navigation assistant. The dataset, termed \textsc{GuideData}, was carefully anonymized, organized for ease of use, and open-sourced via a project webpage to promote human-centered research and inform design decisions in the development of assistive systems for BLV people.

Building on insights from our formative study, we aimed to define an ideal navigation system to support the independent mobility of BLV individuals. We selected a quadruped robot platform to ensure high mobility across diverse terrains, and decided not to rely on GPS or LiDAR. LiDAR is costly in terms of price, computing power, and weight, which shortens operating time and increases overall system cost. It also demonstrates limited robustness under adverse weather conditions (e.g., rain, fog, snow) and when dealing with transparent objects (e.g., glass doors, shop windows, plastic barriers) or reflective surfaces (e.g., water puddles, metallic doors, shiny tiled floors), encountered in everyday environments. In addition, LiDAR-based navigation algorithms typically require dense, prebuilt 3D maps, which demand significant computation and intensive memory for embedded systems~\cite{park2022elasticity}. GPS, meanwhile, is unreliable in urban environments and indoors due to signal degradation near tall buildings as well as its inherent top-down measurement limitations.

Instead of global, metric map-based navigation methods~\cite{kummerle2011g2o,durrant2006simultaneous,cadena2016past}, we propose a navigation framework, titled \textsc{GuideNav}, following a Visual Teach-and-Repeat (VT\&R) system~\cite{furgale2010visual}. In this setup, an operator demonstrates routes by driving the robot using a remote controller, after which the robot autonomously repeats the demonstrated path for a BLV user. This framework aligns closely with how guide dogs are trained and work with their handlers when they are familiarized with the environment. 
Moreover, daily activities typically involve routine routes, and our navigation framework specifically targets these common scenarios as a starting point. During the teach phase, the robot records a stream of images and constructs a topological map (instead of a dense metric map) by saving keyframe images along the path. Using the image-based subgoals (i.e., keyframe images) enables vision-only navigation using a single RGB camera on a standalone system (see Fig.~\ref{fig:teaser}).

While promising, VT\&R involves a fundamental challenge: vulnerability to scene variation. Because the algorithm identifies the robot’s location and selects subgoals by comparing observed scenes to saved keyframes, environmental changes (e.g., lighting, object relocation) can drastically reduce retrieval accuracy, often causing the robot to lose its path~\cite{furgale2010visual}. To mitigate this, we adopt the strategy used in \cite{camara2020accurate,suomela2024placenav}: leveraging visual place recognition (VPR) embeddings~\cite{lowry2016visual,berton2022rethinking} instead of raw image pixel matching~\cite{dall2021fast}. VPR descriptors improve robustness to scene variation by incorporating semantic understanding learned from internet-scale datasets. In addition, we integrate Reloc3r~\cite{dong2025reloc3r}, a deep neural network for camera pose estimation, to compute the robot’s position and orientation relative to target keyframes. These estimates are then used to compute low-level control commands (i.e., linear and angular velocities). We found that Reloc3r outperforms dense feature-matching approaches~\cite{sun2021loftr,Edstedt_2024_CVPR,leroy2024grounding}, which often suffer from degraded accuracy when depth information is unreliable (common in outdoor environments) or fail under scene variations.

Field tests of our robot demonstrate that our system can achieve kilometer-scale, long-range autonomous navigation, surpassing prior guide robot demonstrations that relied on more complex sensing and mapping~\cite{cai2024navigating,takagi2025field}. \textsc{GuideNav} requires no metric maps, GPS, LiDAR, or depth sensing, keeping the platform lightweight and low-cost. By chaining short, visually guided segments, \textsc{GuideNav} avoids drift accumulation and adapts to environmental changes, maintaining robustness even with modest computing resources—an important factor for extended operating time.

In user studies, \textsc{GuideNav} successfully guided BLV individuals along multiple outdoor routes, including one measuring 2.02 km round trip. To our knowledge, this represents the first demonstration of kilometer-scale outdoor autonomous guiding for BLV participants using a camera-only system. Participants described the experience as easy to use, while professional trainers noted both its reliability and differences from guide dogs. Workload and usability assessments confirmed low cognitive demand and high satisfaction for most users, while also pointing to areas for refinement and potential applications. 
%
%
In summary, our contributions are threefold:  
\begin{enumerate}[leftmargin=11pt]
\item A large open-source dataset containing rich information about guide dog–handler interactions, real-world navigation challenges, and daily activities of BLV travelers in navigational contexts.
\item A lightweight, visual teach-and-repeat framework that eliminates reliance on GPS, LiDAR, and depth sensing.
\item We present the first real-world validation of kilometer-scale, vision-only autonomous navigation with BLV participants on their daily routes, measuring turning success rates, interventions, and collisions, and demonstrating robust long-range performance under diverse outdoor conditions.
\end{enumerate}

\section{Related Works}

\subsection{Existing Technological Solutions}

\textit{Un-actuated systems} have been designed in various form factors, such as smart canes~\cite{smartcane, slade2021multimodal,chuang2018deep, ye2016co, ulrich2001guidecane, takizawa2015kinect, takizawa2012kinect, faria2010electronic, saaid2016smart} and wearable or handheld devices~\cite{belt, headmount, hirose2018gonet, zeng2017camera, katzschmann2018safe, li2016isana, rodriguez2012assisting, strumillo2018different, dakopoulos2009wearable, ranganeni2023exploring}. While these devices are lightweight and compact, they typically offer only limited navigation assistance---for example, notifying users of nearby obstacles or local paths, but no goal-directed navigation. More critically, the use of acoustic or vibrotactile feedback (rather than physical pulling or pushing) often causes distraction and mental fatigue, raising serious safety concerns, as BLV individuals rely heavily on environmental sounds for localization and situational awareness~\cite{bradley2002investigating}.

In contrast, \textit{actuated systems} can assist BLV individuals in a manner more similar to sighted guides or animal guide dogs. Such systems can enable smooth collision avoidance, higher walking speeds, and shorter travel distances~\cite{clark1986efficiency,miner2001experience, cai2024navigating}. Since the mid-1970s, wheeled mobile robots have been the most widely studied form factor for actuated systems~\cite{Tachi, meldog, melvin2009rovi, tobita2018structure, guerreiro2019cabot, tobita2017examination, kulyukin2004robotic, megalingam2019autonomous, nanavati2018coupled, galatas2011eyedog, kayukawa2019bbeep, takagi2025field, kuribayashi2025wanderguide}. However, wheeled systems face a critical limitation: they cannot reliably handle uneven terrains such as stairs or curbs~\cite{takagi2025field}, making them impractical for real-world outdoor navigation where assistance is most needed.

Recent advances in quadruped robots have opened new opportunities for mobility assistance. Unlike wheeled systems, quadrupeds provide natural and efficient locomotion across uneven terrains and heterogeneous environments~\cite{xiao2021robotic, chen2023quadruped, hamed2019hierarchical, NSK}. However, existing studies often give limited attention to users’ needs, largely due to a lack of understanding of how guide dogs and their handlers work. For instance, \cite{xiao2021robotic, chen2023quadruped} explored the use of a soft leash to pull a person, whereas both our formative study and prior literature~\cite{hwang2023system, hwang2024towards, guerreiro2019cabot} emphasize that handlers require a rigid harness handle to perceive immediate feedback on the guide dog’s motion for safety. Similarly, some studies positioned the handler behind the robot during navigation~\cite{hamed2019hierarchical, cai2024navigating}, whereas in practice guide-dog handlers walk beside and one step behind the dog’s front feet, enabling them to assess the environment when the dog stops for safety~\cite{hwang2024towards}.

In summary, quadruped robots hold substantial potential compared to other assistive platforms, yet current guide-dog robot research lacks a comprehensive user-centered perspective. Our objective is to develop a pragmatic guide-dog robot that matches—and ultimately surpasses—the abilities of animal guide dogs, providing BLV individuals with a reliable option for independent navigation.

\subsection{Vision-based Navigation and VT\&R}
While prior VT\&R studies demonstrated kilometer-scale path following~\cite{furgale2010visual}, they often relied on precise metric localization or expensive sensors. For example, many VT\&R variants used LiDAR to register the robot’s pose~\cite{furgale2010visual}. More recently, there has been a major departure from such approaches, with new methods aiming to be lightweight and vision-centric. Fischer et al.~\cite{dall2021fast} proposed a bio-inspired scheme that leverages odometry and minimal visual correction (sparse cues) to navigate reliably with low computational cost. However, methods employing naive template matching (e.g., Normalized Cross Correlation~\cite{lewis2001fast}) are fragile to visually similar patterns and often produce false matches under appearance ambiguities.

FFI-VTR~\cite{wang2025ffivtr} advanced this line of work by constructing a topological keyframe graph of feature descriptors, minimizing pixel-level feature distance between matches without requiring global localization. PlaceNav~\cite{suomela2024placenav} further reframed VT\&R subgoal selection as an image-retrieval task: it uses VPR embeddings (CosPlace~\cite{berton2022rethinking}) to pick the next topological subgoal and applies Bayesian filtering for temporal consistency~\cite{xu2021probabilistic}.
While PlaceNav shares some similarity with our GuideNav in subgoal selection, GuideNav also employs a trained network for pose estimation, which enhances robustness to scene and lighting variations and enables reliable long-range navigation.

\section{Formative Study Dataset}

Existing qualitative research spans guide dog ownership ~\cite{whitmarsh2005benefits, refson1999health}, interaction ~\cite{Craigon2017-og, 10.1080/11762322.2010.523626, zhang2023follower}, navigation and communication ~\cite{Gaunet2019-gu, lloyd2008guide2, slade2021multimodal, 10.1145/2513383.2513449}; in parallel, developers of quadruped and other mobility assistive systems report insights from participatory design ~\cite{kuribayashi2023pathfinder, cai2024navigating}, experiments~\cite{liu2024dragon, kim2023transforming, defazio2023seeingeye, wang2023can}, and observations~\cite{hwang2023system, Due02012023, kayukawa2019bbeep}. Few prior studies have released in-depth demonstration videos~\cite{liu2024dragon} or human-robot interaction data~\cite{kim2023transforming, cai2024navigating}; however, there remains a significant gap in large-scale, open-source qualitative resources, especially those documenting handler–guide dog observations or detailed trainer/handler interviews~\cite{xiao2021robotic,hamed2019hierarchical,chen2023quadruped}. \par

To address this gap and inform the human-centered design of \textsc{GuideNav}, we conducted a formative study documenting interviews, navigation, video observations, the matching process, and blindfolded walking sessions. Our de-identified and curated dataset, termed \textsc{GuideData}, comprising 39 interview transcripts, 31 images, and 126 videos with a total duration of over 15 hours, has been open-sourced via our project webpage.

\subsection{Procedure}
The dataset includes detailed documentation for each data type and is composed of five key elements:

\begin{itemize}[leftmargin=11pt]
\item \textbf{Interviews with guide dog handlers:} Semi-structured Q\&A sessions on how they use guide dogs in daily life (transcripts).
\item \textbf{Interviews with guide dog trainers:} Semi-structured Q\&A sessions focusing on how guide dogs are trained and deployed to BLV individuals (transcripts).
\item \textbf{Observation sessions of guide dog–assisted navigation:} Videos capturing how handlers and their guide dogs navigate through environments they walk in regularly (videos).
\item \textbf{Observation sessions of matching training:} Videos documenting how an experienced handler and a new guide dog undergo the initial familiarization process (videos).
\item \textbf{Blindfolded walking sessions by the authors:} Under a guide-dog trainer’s supervision, we conducted blindfolded walks with trained guide dogs to experience handler–dog interaction and the navigational cues needed for safe travel. The video recordings document dog-provided support (e.g., locating sidewalks) versus decisions requiring human judgment (e.g., assessing street-crossing safety), yielding rich evidence on how animal guide dogs deliver navigation assistance.
\end{itemize}
%
\subsection{Participants}
\textsc{GuideData} includes semi-structured interviews with 11 male and 21 female BLV individuals, and observation sessions involving seven guide dog users and one cane user. Participants had an average age of 64.63 years and roughly 24.21 years of experience using navigation aids, with the majority having experience with multiple guide dogs. The dataset also contains interviews, and observation sessions with nine professional guide dog trainers and one O\&M specialist. \par
To ensure the dataset is informative while protecting privacy, we report key demographic attributes such as gender, age, and experience, while anonymizing participant identities. Further demographic details are available in Supplementary Material~1.

\subsection{Data Postprocessing}
To protect privacy, all video recordings are post-processed to blur participants’ faces. For observation videos and images, we applied an automatic face anonymization algorithm~\cite{xu2020centerface,drawitsch2020deface} to obscure participants and visible license plates. All frames were carefully reviewed to correct any failures of the automated process using a video editing tool. For interview transcripts, we removed all personally identifiable information. The resulting dataset is available on Kaggle under a CC0 license for unrestricted public use. Furthermore, we ensured that any author-identifiable details were anonymized in the dataset, webpage, and Kaggle repository. To our knowledge, this is the first comprehensive dataset of interviews and observational recordings that captures rich information on how guide dogs provide assistance in practice. This dataset establishes a strong foundation for defining the specifications of robotic guide dogs and guiding the development of assistive navigation systems.

\section{Technical Requirements for Guide Robots}
Designing a practical guide robot for blind and low-vision travelers is uniquely challenging. Relatively low employment rates demand cost-sensitive solutions, yet the robot must reliably handle diverse real-world environments encountered in daily travel. Grounded in our \textsc{GuideData}, we highlight the critical hardware and navigation requirements that define the specifications of a deployable guide robot. While many dimensions are relevant, we focus on hardware and navigation as foundational; other aspects such as communication, personalization, or user interfaces~\cite{hwang2024towards} remain also important but are deferred for future work.

\subsection{Hardware Constraints}

\subsubsection{Runtime and responsiveness.} The system must operate for at least two hours~\cite{hwang2024towards} at normal walking speeds (up to at least $1.4\,\text{m/s}$)~\cite{bohannon2011normal}, with all perception and control onboard to avoid network dependence. We propose that end-to-end (perception to control) latency must remain under $100\,\text{ms}$ so that safety-critical commands (e.g., ``stop,'') are executed promptly and trusted by the handler. Memory must support storage of large-scale maps. 


\subsubsection{Form factor and weight.} The robot must pass through narrow doors, fit under seats, and be light enough for one person to lift when necessary~\cite{weston2020one}, yet heavy and stable enough to safely halt an adult in emergencies. Legged or wheel–legged locomotion is preferred to negotiate curbs, stairs, and uneven terrain that are encountered in daily travels.

\subsubsection{Minimal Sensor Suite: Vision-Only Approach}
A forward-facing RGB camera is essential for capturing rich visual cues. In contrast, LiDAR – though common in guide robot research – adds cost, power draw, and vulnerability to weather (e.g., drizzle, fog, snow) that corrupt point-cloud returns which can be problematic as handlers may get caught in rain and snow (GH03, GH12). Handlers reported accidents caused by transparent hazards, breaking one’s own leg after falling from ice, or glass doors, which LiDAR often fails to detect. Similarly, GPS proved unreliable in dense urban areas where several participants reported not using GPS-based apps due to frequent signal loss. These findings underscore the rationale for a vision-only approach, as the most practical and scalable option for daily navigation.

\begin{figure*}
  \includegraphics[width=\linewidth]{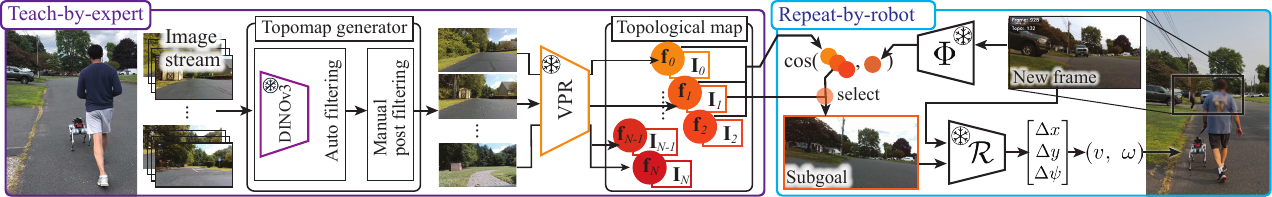}
  \caption{\textsc{GuideNav} teach-and-repeat framework. In the \textit{teach} phase, an expert performs remote control of the robot along the desired route. The robot records a sequence of images, and the topomap generator constructs a topological map by extracting keyframe images. In the \textit{repeat} phase, the robot performs visual place recognition by matching current image features, $\mathbf{f}_t$, extracted by the feature extractor $\Phi$, with the topomap using cosine similarity, and then selects a subgoal. The pose estimator, $\mathcal{R}$, computes the relative pose between the current and subgoal images. \textsc{GuideNav} achieved kilometer-scale autonomous navigation under varying scene conditions. The system was also used by three guide dog handlers and one guide dog trainer, who provided positive feedback on its potential to support mobility assistance for BLV individuals.}
  \Description{System architecture diagram with two phases. Teach-by-expert phase (left): Person walking with guide dog, image stream fed into topomap generator containing DINOv3 auto filtering and manual map design modules, producing topological map with nodes and edges. Repeat-by-robot phase (right): During the robot's navigation beside the human, it performs visual place recognition by computing cosine similarity between current subgoal images and new frame features, selecting matching nodes. Feature extractor Phi processes current view or new frames, pose estimator R computes relative pose changes (delta v, delta theta, delta w) between current and target views.}
  \label{fig:guidnav}
\end{figure*}

\subsection{Navigation Requirements for Guide Robots}



Navigation for BLV travelers imposes requirements distinct from conventional mobile robotics. A guide robot must compactly encode long routes across varied terrains, adhere to orientation principles emphasized in mobility training, and reliably negotiate the diverse obstacles encountered in sidewalk environments. These requirements motivate lightweight, robust approaches such as Visual Teach and Repeat (VT\&R)~\cite{furgale2010visual}, where routes are compactly stored and repeated under changing conditions.

\subsubsection{Mapping and Perception}
Guide robots must represent kilometer-scale routes spanning both structured indoor corridors and unstructured outdoor paths (e.g., cracked sidewalks, curbs, stairs). Compact map representations (e.g., visual descriptors) are preferable to dense 3D point clouds for efficiency. Perception must remain reliable across lighting, weather, and seasonal changes, including transitions from daylight to low-light operation.

\subsubsection{Domain Knowledge for BLV Travel}
Navigation must integrate orientation and mobility (O\&M) practices. For instance, before street crossings the robot should stop at curb edges within one stride, allowing the traveler to probe with their foot and decide when to proceed. O\&M training emphasizes maintaining straight-line travel and executing turns only at right angles on command. In areas without sidewalks, teams often ``shoreline’’ the left road edge to preserve orientation – patterns the robot should replicate.

\subsubsection{Obstacle Avoidance}
Trainers specify that safe clearance requires at least $0.9\,\text{m}$ laterally and $1.8\,\text{m}$ vertically. Beyond standard obstacles, guide robots must detect overhanging (branches, signs, wires) and transparent hazards (glass panes, puddles, ice) that often challenge vision systems. Handler interviews also highlighted everyday obstructions – manholes, construction equipment, and parked cars – that block pedestrian paths and must be negotiated.

\section{Proposed \textsc{GuideNav} Architecture}
Unlike monolithic end-to-end policies, our approach integrates modular components---topological mapping, visual place recognition, relative pose regression, and control---to achieve kilometer-scale outdoor route following on embedded hardware.

\subsection{Problem Formulation}
We define the guide dog route following problem as enabling a robot to safely repeat expert-taught pedestrian routes using only onboard sensing and compute. Formally, the robot state lies in $SE(2)$, $x_t = (p_t, \psi_t)$, where $p_t \in \mathbb{R}^2$ is position and $\psi_t$ is orientation. At each timestep, the robot observes $I_t$ through a camera.  

During the \textbf{teach phase}, an expert demonstrates a safe route $\Gamma = [I_1, \dots, I_N]$. From the demonstration, we construct a sparse topological map $\mathcal{M} = \left\{(I_i, \mathbf{f}_i)\right\}_{i=1}^N$, where $I_i \in \mathbb{R}^{H \times W \times 3}$ are keyframes and $\mathbf{f}_i$ are visual embeddings extracted from a place recognition encoder. The keyframes are extracted using our topological map generator along with a manual filtering process detailed in Section~\ref{subsec:topo}.

During the \textbf{repeat phase}, the estimates the most likely subgoal $I_{g_i} \in \mathcal{M}$ and estimates the relative transformation $\xi_{t \to g_i} \in se(2)$. This relative pose is converted into velocity commands $(v_t, w_t)$ under a nonlinear feedback controller, producing safe closed-loop motion along the expert-taught route.  


\textsc{GuideNav} functions more like a global planner, autonomously following expert-taught subgoals without GPS or dense metric maps. By design, the system does not include a dedicated local planner, but its modular architecture allows one to be seamlessly integrated for detours or dynamic avoidance when required. This balance of simplicity and modularity ensures both efficiency on embedded hardware and adaptability for future extensions.

\subsection{Topological map generator}
\label{subsec:topo}

Each collected frame $I_t$ is encoded using the DINOv3 foundation model~\cite{simeoni2025dinov3} $\mathcal{E}$ into an $\ell_2$-normalized global descriptor $\mathbf{z}_t=\mathcal{E}(I_t)$. We select keyframes with an \emph{adaptive selector} that jointly enforces (1)~minimum temporal spacing, (2)~appearance diversity over a recent buffer via cosine similarity, and (3)~adaptive similarity thresholds that adjust based on trajectory progress. This embedding-driven approach yields compact yet discriminative topological maps, retaining only a small fraction of frames while improving robustness under viewpoint/illumination changes and reducing memory footprint (e.g., $\sim$24~MB for 1~km travel). A human operator may filter low-quality keyframes (e.g., motion-blurred), though this step could be automated in future work.

\subsection{Visual place recognition with temporal consistency}
Each incoming image $I_t$ is encoded by CosPlace~\cite{berton2022rethinking} into a normalized descriptor $\mathbf{f}_t = \Phi(I_t;\theta) \in \mathbb{R}^{512}$. Cosine similarity enables efficient retrieval of candidate subgoals. To prevent erratic frame switching, we enforce temporal consistency through a belief update over topological states, following the approach of Suomela et al.~\cite{suomela2024placenav}. Based on observation, this yields smoother subgoal selection and mitigates the kidnapping problem~\cite{thrun2005probabilistic}.

\subsection{Direct relative pose estimation}
Given a current frame $I_t$ and retrieved subgoal $I_{g_i}$, we estimate their relative pose using ReLoc3r~\cite{dong2025reloc3r} as our pose regression network $\mathcal{R}$:
\begin{equation}
\mathcal{R}(I_t, I_{g_i}; \Theta) = T_{t \to g_i} \in SE(3),
\end{equation}
with $T_{t \to g_i } = \begin{bmatrix} R & \mathbf{t} \\ \mathbf{0}^\top & 1 \end{bmatrix}$.  
The transform is projected onto the ground plane to yield
\begin{equation}
\xi = [\Delta x, \Delta y, \Delta \psi]^\top \in \mathbb{R}^3 \end{equation}
a compact 2D displacement and yaw representation directly consumed by the controller. Rather than relying on Perspective-n-Point (PnP)~\cite{fischler1981random}, which requires dense feature matching and is vulnerable to low texture and depth errors, using a direct regression approach provides robust pose estimates suitable for long-range navigation.

\begin{figure*}[h]

\includegraphics[width=\linewidth]{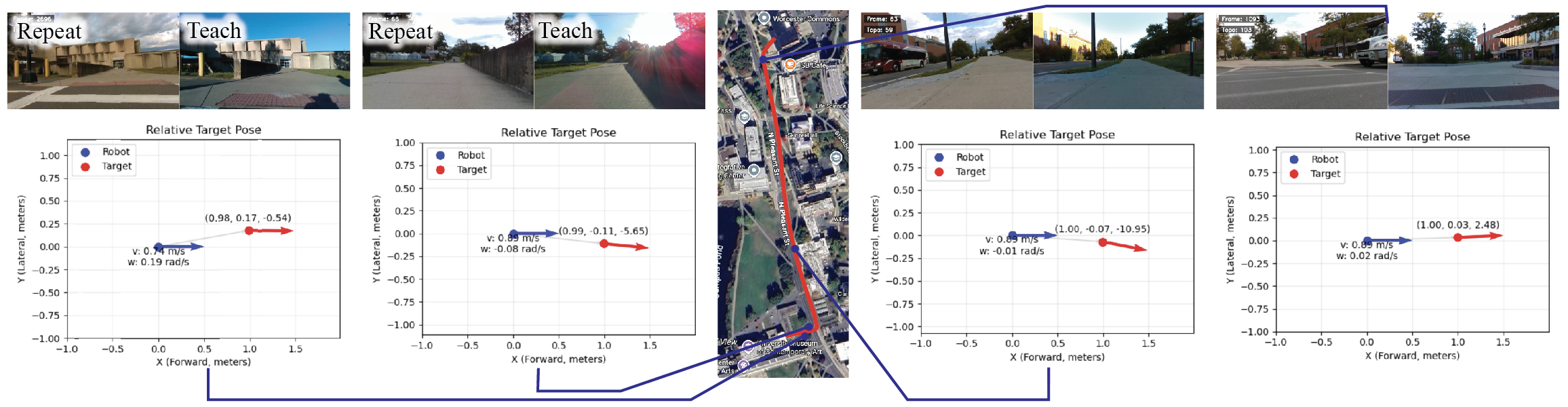}
\caption{\textbf{\textsc{GuideNav} outputs during guidance of H02, showing topological map node prediction and relative pose estimation. The system correctly predicts the node despite drastic lighting changes and unseen objects, and maintains accurate pose estimation.}}
\label{f3:run}
    \Description{Four pairs of teach-repeat image comparisons with central aerial map showing red route path. Each pair contains teach and repeat phase street-level images above scatter plots showing relative target pose with robot and target positions. Second pair demonstrates system robustness with dramatic pink sunset lighting in teach phase versus bright daylight in repeat phase. Plots show consistently accurate forward distance around one meter with small lateral and angular variations across all locations.}
\end{figure*}

\begin{table}
  \caption{User Study Participant Demographics}
  \label{tab:participants2}
  \resizebox{\linewidth}{!}{
  \begin{tabular}{cccccc}
    \toprule
    \textbf{ID} & \textbf{Age} & \textbf{Gender} & \textbf{Vision Level} & \textbf{Experience$^{*}$} & \textbf{Role} \\
    \midrule
    H01 & 63 & F & Totally blind & 35 & User  \\ 
    H02 & 66 & F & Legally blind & 9 & User   \\ 
    H03 & 67 & M & Totally blind & 15  & User \\ 
    \midrule
      T01 & 37 & F &  & 10  & Evaluator   \\ 
    T02 & 35 & F &  & 4  & User   \\ 
    T03 & 28 & F &  & 7  & Evaluator   \\ 

    \bottomrule
\end{tabular}
}
\footnotesize{\\$^{*}$ Indicates the number of total years the participant trained guide dogs.}
\end{table}

\subsection{Goal Position to Velocity Commands}
We model the robot with unicycle kinematics, where the relative pose $\xi = [\Delta x, \Delta y, \Delta \psi]^\top$ is expressed in polar form:
\begin{align}
\rho &= \sqrt{{\Delta x}^2 + {\Delta y}^2}, \quad
\alpha = \arctan2(\Delta y, \Delta x), \quad
\beta = \psi - \alpha,
\end{align}
with $\rho$ the goal distance, $\alpha$ the heading error, and $\beta$ the orientation misalignment. A nonlinear feedback law regulates these terms:
\begin{align}
v_{\mathrm{raw}} &= v_{\max}\!\left(1 - e^{-k_\rho \rho}\right), \\
w_{\mathrm{raw}} &= k_\alpha \alpha + k_\beta \beta,
\end{align}
where $(k_\rho, k_\alpha, k_\beta)$ are control gains. For stability, we further apply three heuristics: reduced speed near the goal, velocity damping under large heading errors, and amplified orientation correction in final alignment. Finally, coordinated scaling enforces velocity limits, yielding smooth, bounded trajectories consistent with unicycle dynamics and reliable teach-and-repeat navigation.


\section{Evaluation of \textsc{GuideNav} Components}

We evaluated \textsc{GuideNav} in both indoor and outdoor environments to assess the robustness of its key components. As shown in Fig.~\ref{f3:run}, the topological map maintained reliable node prediction under lighting changes and partial occlusions.  

\subsection{Topological Map Node Prediction}
We first examined the effect of different keyframe selection strategies on route following. Specifically, we compared a simple distance-based heuristic with topological maps constructed from visual embeddings. Learning-based embeddings consistently improved repeat reliability over the distance heuristic, with DINOv3 providing the most robust localization across varied lighting and viewpoint conditions.  

\subsection{Relative Pose Estimation}
We next compared direct relative camera pose regression with a feature-matching + PnP pipeline. While dense matching can yield accurate correspondences, it frequently failed under textureless surfaces and extremely bright conditions. In contrast, direct regression achieved higher stability, lower failure rates, and smoother control signals, making it better suited for long-horizon outdoor navigation.  





\section{Full System Evaluation with Stakeholders}


We evaluated our full system in realistic contexts through two complementary studies. First, we conducted a field study with three BLV individuals, all experienced guide dog handlers. Second, we replicated an official guide dog evaluation by testing the robot in a standard training environment: a blindfolded guide dog trainer served as the user, while another trainer acted as evaluator. Our objectives were twofold: (1) to assess the effectiveness and acceptance of the guide dog robot among BLV users, and (2) to benchmark performance against the criteria applied to animal guide dogs for real-world deployment.  

\begin{table*}[t]
  \centering
  \caption{Robot Solo and Handler-Guide Dog Robot Team Navigation Results}
  \label{tab:navresults}
  \begin{tabular}{l cc ccc}
    \toprule
    \textbf{Trajectory ID} & \multicolumn{2}{c}{\textbf{Environment}} & \multicolumn{3}{c}{\textbf{Performance}} \\
    \cmidrule(lr){2-3} \cmidrule(lr){4-6}
    & \textbf{\# TSR} & \textbf{Distance (m)} & \textbf{\# Collisions} & \textbf{\# Interventions} & \textbf{Time} \\
    \midrule
    Robot Solo$ \; (\text{FidelcoFront} \rightarrow \text{Storage})$ & 5/6 & $130\,\si{\meter}$ & 0 & 2 & \SI{4}{\minute}\,\SI{8}{\second} \\
    Robot Solo$ \; (\text{FidelcoFront} \rightarrow \text{Parking})$ & 6/7 & $160\,\si{\meter}$ & 0 & 1 & \SI{4}{\minute}\,\SI{2}{\second}\\
    Robot Solo$ \; (\text{LongWalkLoop})$ & 17/17 & $1670\,\si{\meter}$ & 0 & 1$^{\dagger \ddagger}$ & \SI{44}{\minute}\,\SI{58}{\second} \\
    Robot Solo$ \; (\text{ResidentialLoop})$ & 13/14 & $430\,\si{\meter}$ & 1 & 3 &  \SI{31}{\minute}\,\SI{12}{\second} \\
    Robot Solo$ \; (\text{ClassroomLoop--Indoor})$ & 4/6 & $30\,\si{\meter}$ & 0 & 3$^{*}$ & \SI{2}{\minute}\,\SI{42}{\second} \\
    \midrule
    \textbf{Total} & 45/50 (90\%)  & $2420\,\si{\meter}$ & 1 & 10 & \SI{87}{\minute}\,\SI{2}{\second}  \\
    \midrule
    H01$+ \text{Robot} \; (\text{Lot} \rightarrow \text{Library})$ & 4/4 & $390\,\si{\meter}$ & 0 & 0 & \SI{12}{\minute}\,\SI{14}{\second} \\
    H01$+ \text{Robot} \; (\text{Library} \rightarrow \text{Lot})$ & 3/3 & $390\,\si{\meter}$ & 0 & 0$^\dagger$ & \SI{12}{\minute}\,\SI{54}{\second} \\
    H02$+ \text{Robot} \; (\text{Home} \rightarrow \text{BusStop})$ & 7/7 & $70\,\si{\meter}$ & 0 & 0 & \SI{3}{\minute}\,\SI{10}{\second} \\
    H02$+ \text{Robot} \; (\text{Dining} \rightarrow \text{FineArts})$ & 3/3 & $670\,\si{\meter}$ & 0 & 0 & \SI{13}{\minute}\,\SI{43}{\second} \\
    H02$+ \text{Robot} \; (\text{FineArts} \rightarrow \text{Dining})$ & 2/2 & $670\,\si{\meter}$ & 0 & 1$^{\dag}$ & \SI{18}{\minute}\,\SI{6}{\second} \\
    H03$+ \text{Robot} \; (\text{Home} \rightarrow \text{Church})$ &  8/8 & $1010\,\si{\meter}$ & 0 & 1$^{\ddagger}$ & \SI{21}{\minute}\,\SI{5}{\second} \\
    H03$+ \text{Robot} \; (\text{Church} \rightarrow \text{Home})$ & 8/8 & $1010\,\si{\meter}$ & 0 & 3$^{\ddagger}$ & \SI{24}{\minute}\,\SI{3}{\second} \\
    T02$+ \text{Robot} \; (\text{Guide dog evaluation loop})$ & 11/11 & $730\,\si{\meter}$ & 1 & 1$^{\ddagger}$ & \SI{18}{\minute}\,\SI{57}{\second} \\
    \midrule
    \textbf{Total} & 46/46 (100\%)  & $4940\,\si{\meter}$ & 1 & 6 & \SI{124}{\minute}\,\SI{12}{\second}  \\
    \bottomrule
  \end{tabular}
  \footnotesize{\\$^{\dag}$ Technical malfunction (camera disconnection/overheating) - excluded from intervention count; $^{\ddagger}$ Single guidance intervention required for unexpected obstacle avoidance; \\ $^{*}$ Single physical intervention required to remove entangled cables from robot leg; TSR stands for Turning Success Rate}

\end{table*}

\subsection{Participants}



We recruited three experienced guide dog handlers, each capable of independently navigating their local communities without a sighted guide. All relied primarily on a guide dog as their mobility aid. Eligibility criteria were (1) visual acuity at or below the legal blindness threshold defined in the Social Security Act \S 1614~\cite{ssa20}, and (2) at least six months of experience working with a guide dog. Demographic details are shown in Table~\ref{tab:participants2}. H01 and H03 were totally blind (with H03 retaining light perception), while H02 had residual vision (20/2200 in one eye, lower in the other). In addition, three professional guide dog trainers were recruited, each with a minimum of four years’ experience. T01 observed the robot guiding a handler and provided evaluation as an expert bystander. T02 and T03 jointly completed the standard guide dog evaluation course, with T02 blindfolded in the role of user and T03 serving as evaluator.

\subsection{Guide Dog Robot}
Our guide dog robot is a compact, stand-alone quadrupedal platform designed for both indoor and outdoor route following. Our system is built on the Unitree Go2 robot equipped with a rigid harness handle that allows a rigid connection with the user. A single Intel RealSense D435i camera is used for \textsc{GuideNav}. Relative pose estimation is performed using Reloc3r-512~\cite{dong2025reloc3r} with mixed-precision inference to achieve a high accuracy and inference speed. \textsc{GuideNav} runs at 5 Hz on a NVIDIA Jetson AGX Orin.

\subsection{Procedure}
The experiment consisted of four phases:

\begin{enumerate}[leftmargin=15pt]
\item \textbf{Guide dog demonstrations:}
We observed that all handlers (H01–H03) walked familiar community routes with their guide dogs. H01 walked from the parking lot drop-off to her workplace $(\text{Lot} \leftrightarrow \text{Library})$. H02 began walking from home to the bus stop $(\text{Home} \rightarrow \text{BusStop})$ but switched to another route $(\text{Dining} \leftrightarrow \text{FineArts})$ to avoid drawing neighbors’ attention. H03 walked a residential loop from home to the end of the sidewalk near a church which was over a kilometer one-way $(\text{Home} \leftrightarrow \text{Church})$. Additionally, trainer T02 walked a standard evaluation route blindfolded under the supervision of evaluator T03.

\item \textbf{\textsc{GuideNav} Teach (Data collection with robot):}
Based on these observations, a researcher teleoperated the robot along each route, with minor adjustments for efficiency and safety (e.g., walking on the edge of a road instead of the center, bypassing unnecessary stops such as poles). The camera mounted on the robot recorded video to construct the topological map.

\item \textbf{\textsc{GuideNav} Repeat (Autonomous guiding trials):}
\textsc{GuideNav} then autonomously guided the users. Prior to each trial, handlers adjusted the harness handle (length and angle) and practiced walking with the robot (all handlers except for the trainer had prior walking experience with similar prototypes). On a single command, the robot started and navigated to the destination. Users were notified just before the start and after the robot stopped when it reached the goal. One researcher logged data using a laptop for debug purposes (does not affect \textsc{GuideNav} performance), and another video-recorded the walk.

\item \textbf{Post-Trial Interviews and Surveys:}
After each walk and a brief break, participants completed a semi-structured interview and workload/usability surveys. Interviews were audio-recorded, transcribed using a speech-to-text service~\cite{rev2024}. We note that large language model–based generative AI tools were used only to improve the clarity and style of writing.


\end{enumerate}



\subsection{Findings}
All participants successfully reached the goal in their familiar daily routes within their local community with the guidance of the guide dog robot. The distance and the complexity of the daily routes varied as detailed in Table~\ref{tab:navresults}. For example, H03 successfully walked with the robot for over $2\,\si{\kilo\meter}$ ($1\,\si{\kilo\meter}$ one-way) where he usually goes for a walk with his guide dog. H02 in the other hand, walked for a short distance ($70\,\si{\meter}$) in her residency area, however, there were multiple angled and sharp turns to complete the path. The data collection time and \textsc{GuideNav} runtime was held on a different day and time, and various changes in the environment (e.g., parked car, a crowd of pedestrians passing by - jogging, riding bicycles, light changes) made VT\&R more challenging due to appearance change. In total, six interventions were required for guiding the user over $4.9\,\si{\kilo\meter}$.

\subsubsection{Satisfying and Easy Outdoor Travel in Familiar Routes}
GuideNav enabled all BLV handlers to successfully complete their familiar community routes, ranging from 70\,m to over 2\,km round trip, with minimal interventions. Participants consistently expressed surprise that the robot was fully autonomous, even through routes with multiple sharp turns. As H01 remarked, \textit{``Really it was a hundred percent itself? Even around plant planters, that was all on its own. Wow. It was a success.''} Similarly, H02 reflected after a repeat trial, \textit{``So it really did know where it was going and we did do the route, wow.''} Walking with the robot was described as straightforward and low effort. H03 summarized, \textit{``It seems like it's got some potential for it, for sure. It wasn't hard or anything. It was pretty easy.''} Both handlers and trainers highlighted the robot’s ability to negotiate uneven terrain. During observation of H03 walking with his guide dog, we noted surface tree roots. And later in the teach phase, we teleoperated the robot to safely maneuver over them, and in the repeat phase, the robot autonomously guided the user across the same spot. This led H03 to remark it was \textit{``pretty seamless,''} and T01 to observe, \textit{``Yeah, it actually avoided [them].''}  
Collectively, participants framed the experience as \textit{``easy to walk with''} and emphasized its significance beyond laboratory trials. As H03 put it, \textit{``I honestly think that either yours or this will be the future of blindness, the mobility of blindness.''} These reactions highlight not only the robot’s technical robustness but also its perceived potential as a practical mobility aid.





\subsubsection{Comparison with Guide Dogs: Trainers’ Perspectives}
Professional trainers evaluated our guide dog robot against standard guide dog assessment criteria, noting both promising capabilities and differences. The blindfolded trainer (T02) described the experience as sturdy and enjoyable, while the evaluator (T03) concluded, \textit{``She looked comfortable. I thought it was very safe in its movements. I didn’t feel like she was going to run into anything.''} An experienced trainer (T01), who had previously seen the robot under manual control, remarked on its progress and usability: \textit{``Compared to two years ago, the technology has come a long way as far as the actual usability of it in real life. I was impressed by its ability to navigate [...] the concept of putting in a route and following it specifically was very impressive.''}  

At the same time, T02 emphasized that the robot’s movement remained fundamentally different from that of a guide dog, describing it as more abrupt and less fluid: \textit{``To put it very basically, the movement of the robot is just very robotic. It doesn't have the organic flow that a dog does. Even when you're approaching a curve with a dog, you can feel through the harness that the dog is slowing. With the robot, you get fewer of those body cues because the robot doesn’t have them.''}

\subsubsection{Workload and Usability Evaluation}

NASA–TLX instrument~\cite{HART1988139} results showed a clear divide between BLV handlers and professional trainers (Table~\ref{tab:workload_usability}). All three BLV handlers (H01--H03) reported very low workload (median TLX scores below 6 for mental, physical, and frustration), describing the robot’s guidance as requiring minimal cognitive or physical effort. In contrast, the blindfolded trainer (T02) rated workload much higher (48/100), reflecting both a stricter professional benchmark (guide dog expectations) and first-time use of the robot. One handler (H02) showed a large reduction in workload across sessions, with mental demand, effort, and frustration dropping substantially in her second trial, aligning with contextual factors -- her first route was disrupted by neighborhood distractions, although it did not affect \textsc{GuideNav}'s performance.  

System Usability Scale (SUS) scores also varied. Two handlers (H01, H03) rated the system extremely high (92.5 and 95.0), describing it as intuitive and effortless. H02 rated it much lower ($32.5 \rightarrow 40.0$), reflecting initial difficulties and possibly reduced comfort due to her residual vision when told the robot would guide autonomously. The trainer (T02) rated the robot at 52.5, notably lower than her benchmark for a guide dog (80.0), underscoring a usability gap. Overall, BLV users experienced the system as easy and non-frustrating (low TLX, high SUS), while trainers applied stricter standards. These results suggest that \textsc{GuideNav} can achieve very high usability for some users, but evaluations must account for adaptation, personal context, and professional expectations.

\begin{table}[t]
  \centering
  \caption{Perceived workload (raw 0--20 NASA--TLX scores)}
  \label{tab:workload_usability}
\begin{tabular}{lcccc|c}
    \toprule
    \textbf{Component} & \textbf{H01} & \textbf{H02} & \textbf{H03} & \textbf{T02} & \textbf{Median} \\
    \midrule
    Mental demand   & 4  & 5 $\rightarrow$ 4   & 8  & 10 & 6.25 \\
    Physical demand & 1  & 1 $\rightarrow$ 2   & 5  & 10 & 3.25 \\
    Temporal demand & 2  & 0 $\rightarrow$ 0   & 5  & 5  & 3.50 \\
    Performance     & 1  & 0 $\rightarrow$ 5   & 5  & 4  & 3.25 \\
    Effort          & 2  & 5 $\rightarrow$ 2   & 7  & 16 & 5.25 \\
    Frustration     & 1  & 10 $\rightarrow$ 4  & 1  & 3  & 2.00 \\
    \midrule
    \textbf{Total}  & 11 & 21 $\rightarrow$ 17 & 31 & 48 & 25.00 \\
    \bottomrule
  \end{tabular}
\end{table}





\section{Discussions and Conclusion}
Our study shows that \textsc{GuideNav} can already deliver usable guidance in familiar community routes, yet several design refinements remain critical. Both handlers and trainers emphasized the need for smoother motion cues and explicit landmark indication (e.g., signaling curbs, crossings, or turns) to better support orientation practices taught in mobility training. Beyond these improvements, trainers envisioned diverse applications: as a lower-maintenance alternative for those unable or unwilling to care for a guide dog; as a training tool for instructors to assess mobility skills, balance, and pace before matching a person with a guide dog; as a temporary substitute when users are between guide dogs (getting a new one); and as a flexible solution for pre-planned routes in settings such as hospitals, conferences, or community centers. Because the system requires only a single camera, it could be embodied in multiple robot forms and deployed at scale, broadening access for both cane users and guide dog users. Together, these findings highlight not only the feasibility of vision-only teach-and-repeat navigation, but also its potential to fill important gaps in assistive mobility, opening new directions for human-centered guide robot design.  
\bibliographystyle{ACM-Reference-Format}
\bibliography{bibliography}







\end{document}